%% file: iclr2026_conference.tex
\documentclass{article} 
\usepackage{iclr2026_conference,times}

\input{math_commands.tex}

\usepackage{hyperref}
\usepackage{url}
\usepackage{graphicx}
\usepackage{amsthm}
\usepackage{amsmath}
\usepackage{amssymb}
\newtheorem{proposition}{Proposition}
\usepackage{booktabs}
\usepackage{siunitx}
\usepackage{multirow}
\usepackage{algorithm}
\usepackage{placeins}
\usepackage{algpseudocode} 
\makeatletter
\newcommand{\AlgComment}[1]{\(\triangleright\)~\textit{#1}}
\makeatother

\title{The Lifecycle Principle: Stabilizing Dynamic Neural Networks with State Memory}

\author{Zichuan Yang\\
School of Mathematical Sciences\\
Tongji University\\
Shanghai, 200092, China \\
\texttt{2153747@tongji.edu.cn}
}

\iclrfinalcopy 
\begin{document}

\maketitle

\begin{abstract}
I investigate a stronger form of regularization by deactivating neurons for extended periods, a departure from the temporary changes of methods like Dropout. However, this long-term dynamism introduces a critical challenge: severe training instability when neurons are revived with random weights. To solve this, I propose the Lifecycle (LC) principle, a regularization mechanism centered on a key innovation: state memory. Instead of re-initializing a revived neuron, my method restores its parameters to their last known effective state. This process preserves learned knowledge and avoids destructive optimization shocks. My theoretical analysis reveals that the LC principle smooths the loss landscape, guiding optimization towards flatter minima associated with better generalization. Experiments on image classification benchmarks demonstrate that my method improves generalization and robustness. Crucially, ablation studies confirm that state memory is essential for achieving these gains.
\end{abstract}

\section{Introduction}
\label{1}

Biological systems, like the human brain, are not static. They constantly adapt through processes like apoptosis, where cells are pruned, and neurogenesis, where new connections are formed. This dynamic nature is crucial for learning and robustness. Inspired by this, I explored how to create a similar dynamic process inside an artificial neural network. My goal was to develop a new form of regularization.

Existing methods like Dropout introduce temporary changes to the network at each training step~\citep{srivastava2014dropout}. This led me to a core question: what if these changes were not temporary, but lasted for longer periods? A mechanism that deactivates neurons for many epochs could force the network to build more robust and redundant representations. This is the central motivation for my work.

However, this idea presents a significant challenge. A long-term deactivation mechanism can severely disrupt the optimization process. If a neuron is deactivated and later re-initialized with random weights, it introduces a large, abrupt shock to the network. My initial experiments confirmed this. Such a process leads to highly unstable training and poor performance. The network struggles to adapt to these sudden, destructive changes. This observation revealed a core problem: long-term dynamism requires a stable revival mechanism.

To solve this stability problem, I propose the \textbf{Lifecycle (LC) principle}. The key component of this principle is \textbf{state memory}. This is my main contribution. Instead of re-initializing a revived neuron, my method restores its weights and bias to the values they had at the moment of deactivation. This memory-aware revival allows the neuron to reintegrate smoothly. It preserves the useful knowledge the neuron had previously learned. This avoids the destructive shock of random re-initialization. I also introduce a warm-up phase to further soften this re-integration.

My theoretical analysis suggests that the LC principle acts as a smoothing regularizer on the loss landscape. This encourages the optimizer to find flatter minima, which are often associated with better generalization~\citep{keskar2017on}. My analysis also connects the LC mechanism to a reduction in model capacity, providing a link to formal generalization bounds.

I evaluate my method on standard image classification benchmarks. I compare it against baseline models and other common regularization techniques. My experimental results show that the Lifecycle principle can improve generalization and robustness. The results also include ablation studies. These studies confirm the critical role of the state memory component for achieving stable and effective training.The remainder of this paper is organized as follows: Section~\ref{2} reviews related work. Section~\ref{3} details the methodology. Section~\ref{4} describes the experimental setup. Section~\ref{5} conducts theoretical analysis, and Section~\ref{6} concludes the paper.

\section{Related Work}
\label{2}
My work is related to several areas of neural network research. These areas include stochastic regularization, network pruning, dynamic sparse training, and network growth. I will discuss each of these areas to precisely position my contribution.

\subsection{Stochastic Regularization}
\label{2.1}
Dropout randomly sets neuron activations to zero during training to prevent co-adaptation~\citep{srivastava2014dropout}. DropConnect is a similar method that sets individual weights to zero~\citep{wan2013regularization}. These methods apply changes that are temporary and memoryless. The network structure is only altered for a single forward and backward pass. My Lifecycle principle is different. It introduces structural changes that are long-term. A neuron remains inactive for many epochs. This provides a different and potentially stronger regularization signal.

\subsection{Network Pruning}
\label{2.2}
Network pruning aims to reduce model size and improve efficiency. Pruning methods permanently remove unimportant weights or neurons from a trained network~\citep{han2015learning, blalock2020what}. The goal of pruning is typically model compression, not regularization. The removal of parameters is permanent. In contrast, my LC mechanism is designed for regularization. The deactivation of neurons is temporary, and they can be revived later in the training process.

\subsection{Dynamic Sparse Training}
\label{2.3}
Dynamic Sparse Training (DST) methods are perhaps the most related to my work. These methods change the network's sparse connectivity during training. For example, SET prunes and randomly adds new connections in each epoch~\citep{mocanu2018scalable}. RigL uses gradient information to decide which new connections to grow~\citep{evci2020rigging}. These methods explore different sparse subnetworks. However, a key difference exists. When DST methods grow new connections, they are typically initialized to zero or small random values. This can introduce instability, as the network must learn these new parameters from scratch. My work directly addresses this specific problem. The Lifecycle principle operates at the neuron level, not the connection level. Most importantly, my core contribution of state memory ensures that a revived neuron is not re-initialized. It is restored to its last known effective state. This design choice is a direct solution to the optimization challenge that arises from abrupt structural changes in dynamic networks.

\subsection{Network Growth and Neurogenesis}
\label{2.4}

Another related area is network growth, sometimes referred to as neurogenesis. These methods start with a small network and dynamically add new neurons or layers during training. For example, Net2Net can instantly make a network wider or deeper while preserving its function~\citep{chen2015net2net}. Other methods, like Dynamically Expandable Networks (DEN) add neurons to adapt to new tasks in continual learning settings~\citep{yoon2018lifelong}. The primary goal of these methods is to find an optimal network architecture or to adapt to new data distributions.

My Lifecycle principle differs from network growth in two fundamental ways. First, the motivation is different. I use dynamic structural changes for regularization within a fixed-capacity architecture, not for finding a better architecture or expanding capacity. Second, and more critically, the mechanism is different. Network growth methods add brand new neurons that are randomly initialized. In contrast, my LC mechanism revives a specific, previously existing neuron. The revival process is guided by state memory, restoring the neuron to a previously useful state. This distinction is central to my work's goal of maintaining training stability while still benefiting from dynamic network structures.

\section{Methodology}
\label{3}
In this section, I first explain the high-level concept of the Lifecycle (LC) principle. Then, I detail its specific implementation in a fully connected layer, which I call Lifecycle Linear (LCL).

\subsection{The Lifecycle Principle}
\label{3.1}
\begin{figure}[ht]
\begin{center}
\includegraphics[width=0.8\linewidth]{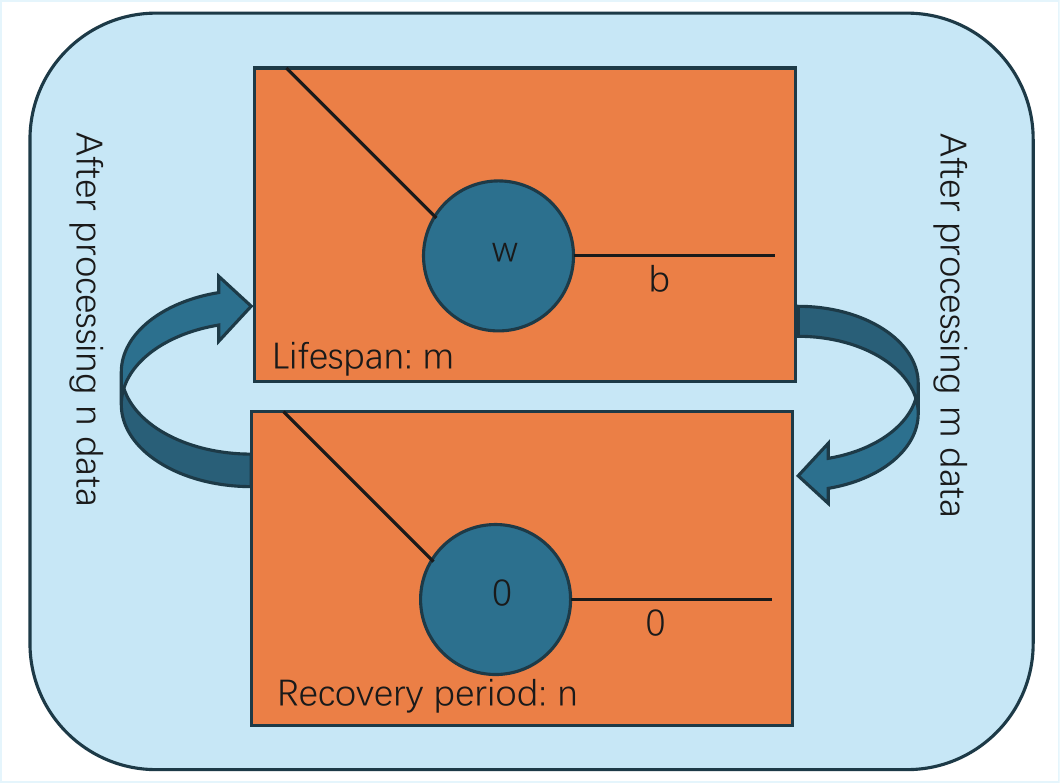}
\end{center}
\caption{Diagram of the three stages of the Lifecycle principle}
\label{fig:LC}
\end{figure}

The Lifecycle principle is a dynamic regularization mechanism. It operates in three main stages for each individual neuron: Deactivation, State Memory, and Revival. The process is vividly shown in Figure~\ref{fig:LC}.

\subsubsection{Deactivation}
\label{3.1.1}
Each neuron in a Lifecycle layer is assigned a random lifespan, measured in training steps. I sample this lifespan from a predefined range. A counter for each neuron tracks its remaining life. When the counter reaches zero, the neuron becomes inactive. It stops contributing to the network's forward and backward passes.

\subsubsection{State Memory}
\label{3.1.2}
This is the core component of my principle. At the exact moment a neuron is deactivated, I save its current weight and bias parameters. These saved parameters are stored in separate, non-trainable buffers. The original parameters of the deactivated neuron remain in the network but do not receive any gradient updates, as their contribution is masked to zero.

\subsubsection{Revival}
\label{3.1.3}
After deactivation, the neuron enters a recovery period. This period is also randomly sampled from a predefined range. When the recovery period ends, the neuron is revived. Upon revival, the neuron does not use random new weights. Instead, it restores the parameters that were saved in its state memory. This memory-aware process allows the neuron to return to a previously useful state.

\subsection{Instantiation on Fully Connected Layers}
\label{3.2}
I implemented the LC principle in a fully connected layer called Lifecycle Linear. This layer maintains several state buffers to manage the lifecycle of each neuron, including counters and the state memory buffers.

The layer uses an "is active" buffer as a binary mask during the forward pass. I multiply the layer's weights and biases by this mask. This effectively removes inactive neurons from the computation graph. As a result, no gradients flow to their parameters, and their weights are frozen.

As I discussed, re-initializing a revived neuron creates a large shock to the optimization. My state memory mechanism directly solves this problem. By restoring the neuron to its last known effective state, the revival process becomes a much smaller, non-destructive perturbation. My ablation studies in the experiments section will show that this memory component is critical for success.

To further smooth the revival process, I introduce a warm-up mechanism. When a neuron is revived, its output contribution is gradually scaled up to its full strength over a predefined number of training steps. I implement this with a "warmup scaler" buffer. This ensures the re-introduction of a neuron is a gradual and stable process. The detailed update steps are provided in Algorithm~\ref{alg:lifecycle-linear-final}.

\section{Experiment}
\label{4}
In this section, I describe the experimental setup I used to evaluate the Lifecycle principle. I detail the datasets, the specific methods I compare, and the evaluation metrics. The goal is to empirically validate the effectiveness of my proposed mechanism.

\subsection{Datasets and Models}
\label{4.1}
I conducted experiments across a range of datasets and model architectures. For training from scratch, I used the CIFAR-10 and CIFAR-100 datasets~\citep{krizhevsky2009learning}. I chose these because they are standard benchmarks for evaluating regularization methods. I used three different model architectures to test for generality. These were ResNet-18~\citep{He2015Deep}, a convolutional network, DeiT-Tiny~\citep{pmlr-v139-touvron21a}, a vision transformer, and EfficientNetV2-S~\citep{tan2021efficientnetv2}, a modern efficient architecture. The results are shown in Table~\ref{tab:cifar10} and Table~\ref{tab:cifar100} respectively.

\subsection{Compared Methods}
\label{4.2}
I compared my method, which I call LCL, against a comprehensive set of baseline and advanced regularization techniques. The Baseline method is a standard training setup without any specific regularization beyond what is in the base architecture. I also compared against weight decay~\citep{krogh1992avoiding}, Dropout~\citep{srivastava2014dropout}, DropConnect~\citep{wan2013regularization}, Scheduled Drop, Label Smoothing~\citep{szegedy2016rethinking}, Stochastic Depth~\citep{huang2016deep}, and RigL~\citep{evci2020rigging}. This wide range of competitors allows for a thorough evaluation of LCL's performance.

To understand the importance of each component of my method, I performed extensive Ablation Studies. I tested variants of LCL where the state memory was replaced with standard re-initialization schemes, which I call LCL-Reinit-He and LCL-Reinit-Xavier. I also tested versions of both the main method and the re-initialization variants without the warm-up mechanism, which I call LCL-no-warmup. These ablations are crucial for demonstrating that state memory is the key to stabilizing the training process.

\subsection{Evaluation Metrics}
\label{4.3}

I used several metrics to evaluate the performance of each method. The primary metric is the final test accuracy (ACC), averaged over multiple runs. To assess robustness, I measured the mean accuracy on corrupted data (mCA) using the CIFAR-C benchmarks~\citep{hendrycks2018benchmarking}. I also measured the Expected Calibration Error (ECE) to evaluate model confidence.

To connect my experimental results with my theoretical analysis, I measured two additional indicators. I estimated the sharpness of the loss minimum by computing the top eigenvalue of the Hessian matrix. I also computed a proxy for model capacity by calculating the product of the spectral norms of the network's layers, which I call the Lipschitz bound.

\begin{table}[htbp]
    \centering
    \caption{Performance comparison of various methods under different architectures on CIFAR-10}
    \label{tab:cifar10}
    
    \sisetup{
        separate-uncertainty,
        table-align-text-post=false
    }
    
    \resizebox{\textwidth}{!}{%

    \begin{tabular}{
        l
        l
        S[table-format=2.2(2)]
        S[table-format=2.2(2)]
        S[table-format=1.4(4)]
        r
        r 
    }
        \toprule
        \textbf{Model} & \textbf{Method} & {\textbf{ACCURACY (\%)}} & {\textbf{mCA (\%)}} & {\textbf{ECE}} & {\textbf{Hessian top eigenvalue}} & {\textbf{Lipschitz bound}} \\
        \midrule
        \multirow{12}{*}{DeiT-Tiny} 
        & Baseline & 82.28 \pm 0.31 & 67.03 \pm 0.70 & 0.1362 \pm 0.0017 & $2.5451 \times 10^{+02} \pm 2.1602 \times 10^{+02}$ & $1.5915 \times 10^{+42} \pm 2.6503 \times 10^{+41}$ \\
        & DropConnect & 77.68 \pm 1.22 & 64.43 \pm 1.24 & 0.1465 \pm 0.0004 & $2.2523 \times 10^{+00} \pm 2.4048 \times 10^{-01}$ & $8.5564 \times 10^{+43} \pm 7.2450 \times 10^{+41}$ \\
        & Dropout & 82.21 \pm 0.71 & 66.82 \pm 0.50 & 0.1387 \pm 0.0046 & $1.5750 \times 10^{-01} \pm 2.6088 \times 10^{-01}$ & $1.1411 \times 10^{+43} \pm 1.9081 \times 10^{+42}$ \\
        & LCL & 83.33 \pm 0.55 & 66.76 \pm 0.83 & 0.1310 \pm 0.0032 & $2.6858 \times 10^{-02} \pm 3.2738 \times 10^{-02}$ & $1.1062 \times 10^{+18} \pm 7.6617 \times 10^{+17}$ \\
        & LCL-Reinit-He & 83.60 \pm 0.05 & 65.90 \pm 0.29 & 0.0805 \pm 0.0017 & $2.2110 \times 10^{+00} \pm 1.4923 \times 10^{+00}$ & $1.6676 \times 10^{+17} \pm 1.9316 \times 10^{+16}$ \\
        & LCL-Reinit-Xavier & 80.29 \pm 0.09 & 66.27 \pm 1.11 & 0.0816 \pm 0.0005 & $2.0244 \times 10^{+00} \pm 4.0499 \times 10^{-01}$ & $1.4175 \pm 10^{+16} \pm 2.3110 \times 10^{+16}$ \\
        & LCL-no-warmup & 79.46 \pm 0.14 & 65.54 \pm 0.24 & 0.0856 \pm 0.0001 & $5.8844 \times 10^{+00} \pm 4.1765 \times 10^{+00}$ & $1.4066 \times 10^{+18} \pm 1.1204 \times 10^{+18}$ \\
        & LCL-Reinit-Xavier-no-warmup & 79.65 \pm 0.58 & 65.95 \pm 0.30 & 0.0578 \pm 0.0185 & $6.5402 \times 10^{+00} \pm 8.8834 \times 10^{+00}$ & $9.8294 \times 10^{+17} \pm 5.1318 \times 10^{+17}$ \\
        & LCL-no-warmup & 82.46 \pm 0.37 & 66.78 \pm 0.24 & 0.1363 \pm 0.0023 & $7.1095 \times 10^{-03} \pm 4.9323 \times 10^{-03}$ & $7.2936 \times 10^{+17} \pm 2.9854 \times 10^{+16}$ \\
        & Label Smoothing & 82.69 \pm 0.23 & 67.39 \pm 0.78 & 0.0451 \pm 0.0023 & $6.3037 \times 10^{-01} \pm 5.3099 \times 10^{-01}$ & $1.0906 \times 10^{+42} \pm 6.0460 \times 10^{+41}$ \\
        & RigL & 80.43 \pm 0.51 & 66.56 \pm 0.62 & 0.1307 \pm 0.0032 & $3.8986 \times 10^{+00} \pm 6.3824 \times 10^{+00}$ & $1.5706 \times 10^{+37} \pm 1.3807 \times 10^{+37}$ \\
        & Scheduled Drop & 81.37 \pm 0.21 & 66.75 \pm 0.34 & 0.0937 \pm 0.0125 & $2.9252 \times 10^{+00} \pm 3.3648 \times 10^{+00}$ & $4.8767 \times 10^{+44} \pm 4.9011 \times 10^{+44}$ \\
        & Stochastic Depth & 82.74 \pm 0.60 & 67.39 \pm 0.49 & 0.1225 \pm 0.0070 & $1.2437 \times 10^{-01} \pm 1.3613 \times 10^{-01}$ & $4.6156 \times 10^{+42} \pm 3.8933 \times 10^{+42}$ \\
        & Weight Decay & 82.51 \pm 0.16 & 67.24 \pm 0.51 & 0.1318 \pm 0.0048 & $2.6994 \times 10^{-01} \pm 1.0701 \times 10^{-01}$ & $1.3318 \times 10^{+42} \pm 1.1462 \times 10^{+42}$ \\
        \midrule
        \multirow{12}{*}{EfficientNetV2-S} 
        & Baseline & 91.75 \pm 0.05 & 72.05 \pm 0.39 & 0.0645 \pm 0.0004 & $6.2794 \times 10^{-01} \pm 1.6461 \times 10^{-01}$ & $7.5175 \times 10^{-22} \pm 9.8524 \times 10^{-22}$ \\
        & DropConnect & 91.95 \pm 0.24 & 70.75 \pm 0.41 & 0.0665 \pm 0.0012 & $3.9553 \times 10^{+01} \pm 8.6014 \times 10^{+01}$ & $1.6440 \times 10^{-20} \pm 2.3193 \times 10^{-20}$ \\
        & Dropout & 91.59 \pm 0.27 & 69.96 \pm 0.69 & 0.0652 \pm 0.0009 & $1.7336 \times 10^{-02} \pm 2.1141 \times 10^{-02}$ & $2.9213 \times 10^{-21} \pm 6.9083 \times 10^{-21}$ \\
        & LCL & 92.27 \pm 0.23 & 70.76 \pm 0.17 & 0.0631 \pm 0.0022 & $1.0896 \times 10^{+02} \pm 1.1036 \times 10^{+02}$ & $3.5958 \times 10^{-23} \pm 6.1396 \times 10^{-23}$ \\
        & LCL-Reinit-He & 91.83 \pm 0.29 & 70.84 \pm 0.30 & 0.0576 \pm 0.0012 & $7.5834 \times 10^{+00} \pm 8.8675 \times 10^{+00}$ & $2.3634 \times 10^{-23} \pm 2.3656 \times 10^{-23}$ \\
        & LCL-Reinit-He-no-warmup & 91.85 \pm 0.11 & 70.67 \pm 0.55 & 0.0587 \pm 0.0032 & $2.4284 \times 10^{+01} \pm 1.4831 \times 10^{+01}$ & $1.0385 \times 10^{-22} \pm 1.5956 \times 10^{-22}$ \\
        & LCL-Reinit-Xavier & 92.18 \pm 0.17 & 71.59 \pm 0.47 & 0.0543 \pm 0.0045 & $4.5209 \times 10^{+01} \pm 1.4300 \times 10^{+01}$ & $1.2075 \times 10^{-23} \pm 1.4580 \times 10^{-23}$ \\
        & LCL-Reinit-Xavier-no-warmup & 91.76 \pm 0.11 & 70.51 \pm 0.57 & 0.0597 \pm 0.0045 & $7.5225 \times 10^{+01} \pm 9.4649 \times 10^{+01}$ & $5.5504 \times 10^{-23} \pm 2.1454 \times 10^{-23}$ \\
        & LCL-no-warmup & 91.99 \pm 0.12 & 70.86 \pm 1.04 & 0.0631 \pm 0.0023 & $2.6111 \times 10^{+01} \pm 4.4439 \times 10^{+01}$ & $3.3241 \times 10^{-23} \pm 4.0754 \times 10^{-23}$ \\
        & Label Smoothing & 91.74 \pm 0.09 & 71.12 \pm 1.03 & 0.0643 \pm 0.0019 & $6.5665 \times 10^{+00} \pm 8.1441 \times 10^{+00}$ & $6.7736 \times 10^{-19} \pm 8.2336 \times 10^{-19}$ \\
        & RigL & 89.52 \pm 0.39 & 68.05 \pm 0.28 & 0.0700 \pm 0.0047 & $1.0623 \times 10^{+02} \pm 1.2399 \times 10^{+02}$ & $6.5993 \times 10^{-11} \pm 9.6405 \times 10^{-11}$ \\
        & Scheduled Drop & 91.68 \pm 0.09 & 70.68 \pm 0.37 & 0.0653 \pm 0.0022 & $1.3636 \times 10^{+00} \pm 2.2324 \times 10^{+00}$ & $1.8530 \times 10^{-21} \pm 1.7848 \times 10^{-21}$ \\
        & Stochastic Depth & 92.40 \pm 0.09 & 71.58 \pm 0.40 & 0.0557 \pm 0.0006 & $1.3775 \times 10^{+01} \pm 2.0836 \times 10^{+01}$ & $2.4586 \times 10^{-21} \pm 3.2792 \times 10^{-21}$ \\
        & Weight Decay & 91.50 \pm 0.34 & 71.22 \pm 1.04 & 0.0631 \pm 0.0022 & $1.0439 \times 10^{-01} \pm 1.8077 \times 10^{-01}$ & $1.4272 \times 10^{-22} \pm 2.0451 \times 10^{-22}$ \\
        \midrule
        \multirow{12}{*}{ResNet-18} 
        & Baseline & 93.43 \pm 0.36 & 75.21 \pm 4.77 & 0.0526 \pm 0.0005 & $4.9701 \times 10^{-02} \pm 4.1458 \times 10^{-02}$ & $2.2214 \times 10^{-01} \pm 1.4510 \times 10^{-01}$ \\
        & DropConnect & 71.89 \pm 17.52 & 60.49 \pm 18.94 & 0.2332 \pm 0.1451 & $1.3182 \times 10^{-01} \pm 2.0798 \times 10^{-01}$ & $4.2358 \times 10^{-01} \pm 7.7402 \times 10^{-01}$ \\
        & Dropout & 93.54 \pm 0.64 & 75.50 \pm 5.04 & 0.0491 \pm 0.0044 & $1.1349 \times 10^{-01} \pm 6.1725 \times 10^{-02}$ & $1.0360 \times 10^{+00} \pm 3.3019 \times 10^{-01}$ \\
        & LCL & 94.19 \pm 0.44 & 76.25 \pm 4.67 & 0.0535 \pm 0.0036 & $3.2529 \times 10^{-02} \pm 6.6334 \times 10^{-02}$ & $1.0329 \times 10^{-01} \pm 1.1494 \times 10^{-01}$ \\
        & LCL-Reinit-He & 93.14 \pm 0.41 & 75.38 \pm 4.14 & 0.0533 \pm 0.0057 & $2.3166 \times 10^{-02} \pm 2.6333 \times 10^{-01}$ & $7.9345 \times 10^{-02} \pm 7.9420 \times 10^{-02}$ \\
        & LCL-Reinit-He-no-warmup & 93.14 \pm 0.60 & 75.15 \pm 4.35 & 0.0505 \pm 0.0069 & $2.6973 \pm 10^{-01} \pm 2.3331 \times 10^{-01}$ & $1.1790 \times 10^{-01} \pm 1.4865 \times 10^{-01}$ \\
        & LCL-Reinit-Xavier & 93.20 \pm 0.59 & 75.24 \pm 3.93 & 0.0507 \pm 0.0103 & $7.7093 \times 10^{-01} \pm 6.4969 \times 10^{-01}$ & $6.5194 \times 10^{-02} \pm 7.2337 \times 10^{-02}$ \\
        & LCL-Reinit-Xavier-no-warmup & 93.26 \pm 0.27 & 75.14 \pm 3.96 & 0.0498 \pm 0.0058 & $5.7553 \times 10^{-01} \pm 9.3336 \times 10^{-01}$ & $9.6476 \times 10^{-02} \pm 9.6587 \times 10^{-02}$ \\
        & LCL-no-warmup & 93.21 \pm 0.73 & 75.54 \pm 4.99 & 0.0552 \pm 0.0055 & $1.0468 \times 10^{-01} \pm 1.4658 \times 10^{-01}$ & $1.1077 \times 10^{-01} \pm 1.4656 \times 10^{-01}$ \\
        & Label Smoothing & 93.50 \pm 0.72 & 75.75 \pm 4.51 & 0.0741 \pm 0.0080 & $5.2853 \times 10^{-03} \pm 3.9109 \times 10^{-00}$ & $3.1636 \times 10^{-01} \pm 1.2444 \times 10^{-01}$ \\
        & RigL & 92.66 \pm 1.37 & 74.33 \pm 2.97 & 0.0578 \pm 0.0099 & $5.6551 \times 10^{-03} \pm 1.8701 \times 10^{-03}$ & $1.3246 \times 10^{+00} \pm 2.0818 \times 10^{+00}$ \\
        & Scheduled Drop & 93.29 \pm 0.71 & 74.88 \pm 4.60 & 0.0627 \pm 0.0057 & $2.0393 \times 10^{-03} \pm 3.4935 \times 10^{-01}$ & $8.6192 \times 10^{-01} \pm 9.2442 \times 10^{-01}$ \\
        & Stochastic Depth & 93.64 \pm 0.62 & 75.31 \pm 4.04 & 0.0528 \pm 0.0029 & $4.2677 \times 10^{-03} \pm 1.3851 \times 10^{-01}$ & $9.5992 \times 10^{-01} \pm 1.9771 \times 10^{-01}$ \\
        & Weight Decay & 93.55 \pm 0.55 & 75.25 \pm 5.04 & 0.0484 \pm 0.0045 & $4.1750 \times 10^{-01} \pm 6.1375 \times 10^{-01}$ & $2.2161 \times 10^{-01} \pm 2.2131 \times 10^{-01}$ \\
        \bottomrule
    \end{tabular}%
    }
\end{table}

\begin{table}[htbp]
    \centering
    \caption{Performance comparison of various methods under different architectures on CIFAR-100}
    \label{tab:cifar100}
    
    \sisetup{
        separate-uncertainty,
        table-align-text-post=false
    }
    
    \resizebox{\textwidth}{!}{%

    \begin{tabular}{
        l
        l
        S[table-format=2.2(2)]
        S[table-format=2.2(2)]
        S[table-format=1.4(4)]
        r 
        r 
    }
        \toprule
        \textbf{Model} & \textbf{Method} & {\textbf{ACCURACY (\%)}} & {\textbf{mCA (\%)}} & {\textbf{ECE}} & {\textbf{Hessian top eigenvalue}} & {\textbf{Lipschitz bound}} \\
        \midrule
        \multirow{12}{*}{DeiT-Tiny} 
        & Baseline & 54.01 \pm 0.58 & 38.67 \pm 0.44 & 0.3379 \pm 0.0026 & $1.3948 \times 10^{-01} \pm 1.9338 \times 10^{-01}$ & $4.7242 \times 10^{+42} \pm 7.1152 \times 10^{+42}$ \\
        & DropConnect & 51.08 \pm 0.69 & 36.95 \pm 0.26 & 0.3262 \pm 0.0065 & $1.7482 \times 10^{-01} \pm 3.2530 \times 10^{-02}$ & $1.9420 \times 10^{+43} \pm 3.2222 \times 10^{+43}$ \\
        & Dropout & 53.59 \pm 0.51 & 38.00 \pm 0.28 & 0.3481 \pm 0.0018 & $2.0531 \times 10^{-03} \pm 3.4067 \times 10^{-03}$ & $1.0061 \times 10^{+43} \pm 2.1348 \times 10^{+43}$ \\
        & LCL & 54.43 \pm 0.24 & 37.72 \pm 0.44 & 0.2744 \pm 0.0324 & $5.5074 \times 10^{-02} \pm 4.8620 \times 10^{-02}$ & $6.9541 \times 10^{+16} \pm 1.4663 \times 10^{+16}$ \\
        & LCL-Reinit-He & 53.83 \pm 0.34 & 38.98 \pm 0.33 & 0.1633 \pm 0.0018 & $1.9300 \times 10^{+00} \pm 2.2615 \times 10^{-01}$ & $1.3616 \times 10^{+16} \pm 2.1533 \times 10^{+15}$ \\
        & LCL-Reinit-Xavier & 52.54 \pm 0.47 & 38.23 \pm 0.14 & 0.1793 \pm 0.0074 & $3.4608 \times 10^{+00} \pm 1.7941 \times 10^{+00}$ & $2.2845 \times 10^{+16} \pm 2.1903 \times 10^{+15}$ \\
        & LCL-no-warmup & 52.33 \pm 0.18 & 37.65 \pm 0.77 & 0.1516 \pm 0.0167 & $7.6328 \times 10^{+00} \pm 4.4879 \times 10^{+00}$ & $1.3006 \times 10^{+17} \pm 3.1906 \times 10^{+17}$ \\
        & LCL-Reinit-Xavier-no-warmup & 52.30 \pm 0.26 & 37.95 \pm 0.25 & 0.1306 \pm 0.0064 & $7.9331 \times 10^{+00} \pm 5.3944 \times 10^{+00}$ & $1.3514 \times 10^{+17} \pm 5.1344 \times 10^{+17}$ \\
        & LCL-no-warmup & 55.11 \pm 0.37 & 38.85 \pm 0.71 & 0.3106 \pm 0.0064 & $2.1349 \times 10^{-02} \pm 3.1156 \times 10^{-02}$ & $1.1329 \times 10^{+17} \pm 6.2936 \times 10^{+16}$ \\
        & Label Smoothing & 53.67 \pm 0.33 & 37.99 \pm 0.18 & 0.1392 \pm 0.0100 & $1.7411 \times 10^{-02} \pm 1.1188 \times 10^{-02}$ & $1.5658 \times 10^{+41} \pm 5.5770 \times 10^{+40}$ \\
        & RigL & 53.20 \pm 0.72 & 37.78 \pm 0.36 & 0.3141 \pm 0.0015 & $6.2725 \times 10^{+00} \pm 1.0343 \times 10^{+00}$ & $1.3380 \times 10^{+40} \pm 1.0807 \times 10^{+40}$ \\
        & Scheduled Drop & 52.37 \pm 0.47 & 37.83 \pm 0.24 & 0.1905 \pm 0.0359 & $9.1753 \times 10^{-01} \pm 1.0371 \times 10^{+00}$ & $3.9858 \times 10^{+43} \pm 2.6709 \times 10^{+42}$ \\
        & Stochastic Depth & 54.24 \pm 0.26 & 39.19 \pm 0.05 & 0.3138 \pm 0.0050 & $1.4653 \times 10^{-02} \pm 2.3980 \times 10^{-02}$ & $2.5946 \times 10^{+42} \pm 3.5641 \times 10^{+42}$ \\
        & Weight Decay & 54.81 \pm 0.41 & 38.48 \pm 0.08 & 0.3321 \pm 0.0025 & $1.2025 \times 10^{-03} \pm 1.8803 \times 10^{-03}$ & $1.6511 \times 10^{+42} \pm 9.2869 \times 10^{+41}$ \\
        \midrule
        \multirow{12}{*}{EfficientNetV2-S} 
        & Baseline & 57.21 \pm 0.73 & 38.56 \pm 1.51 & 0.3018 \pm 0.0020 & $2.8125 \times 10^{-01} \pm 3.2802 \times 10^{-01}$ & $1.0839 \times 10^{-10} \pm 1.1642 \times 10^{-10}$ \\
        & DropConnect & 59.43 \pm 1.89 & 39.36 \pm 1.43 & 0.3107 \pm 0.0110 & $1.1974 \times 10^{+03} \pm 1.7876 \times 10^{+03}$ & $3.6018 \times 10^{-12} \pm 1.3731 \times 10^{-10}$ \\
        & Dropout & 56.54 \pm 0.39 & 37.76 \pm 1.08 & 0.3048 \pm 0.0013 & $2.2120 \times 10^{-02} \pm 2.6759 \times 10^{-02}$ & $3.5782 \times 10^{-11} \pm 4.7974 \times 10^{-11}$ \\
        & LCL & 58.15 \pm 0.23 & 37.18 \pm 2.55 & 0.2645 \pm 0.0464 & $5.3978 \times 10^{-01} \pm 3.5875 \times 10^{-01}$ & $3.6108 \times 10^{-12} \pm 2.3642 \times 10^{-12}$ \\
        & LCL-Reinit-He & 56.45 \pm 0.50 & 37.84 \pm 0.60 & 0.1910 \pm 0.0657 & $4.5453 \times 10^{+02} \pm 7.6228 \times 10^{+02}$ & $1.2214 \times 10^{-11} \pm 1.5044 \times 10^{-11}$ \\
        & LCL-Reinit-He-no-warmup & 56.42 \pm 1.13 & 37.39 \pm 1.01 & 0.1890 \pm 0.0534 & $7.2878 \times 10^{-01} \pm 8.0738 \times 10^{-00}$ & $2.0783 \times 10^{-12} \pm 1.9727 \times 10^{-12}$ \\
        & LCL-Reinit-Xavier & 52.35 \pm 0.41 & 37.17 \pm 0.52 & 0.2212 \pm 0.0477 & $2.1988 \times 10^{+01} \pm 5.1241 \times 10^{+01}$ & $8.8096 \times 10^{-12} \pm 1.1389 \times 10^{-11}$ \\
        & LCL-Reinit-Xavier-no-warmup & 57.35 \pm 0.41 & 37.89 \pm 1.22 & 0.2200 \pm 0.0485 & $2.6606 \times 10^{+02} \pm 6.0844 \times 10^{+01}$ & $5.2928 \times 10^{-12} \pm 5.0175 \times 10^{-12}$ \\
        & LCL-no-warmup & 55.39 \pm 0.86 & 37.09 \pm 0.22 & 0.2296 \pm 0.0906 & $3.1364 \times 10^{+01} \pm 1.8614 \times 10^{+01}$ & $2.8705 \times 10^{-12} \pm 2.6846 \times 10^{-12}$ \\
        & Label Smoothing & 57.45 \pm 0.64 & 38.62 \pm 1.12 & 0.1459 \pm 0.0130 & $2.5083 \times 10^{-02} \pm 2.1937 \times 10^{-02}$ & $2.8466 \times 10^{-12} \pm 9.3446 \times 10^{-13}$ \\
        & RigL & 56.83 \pm 0.28 & 38.94 \pm 0.61 & 0.2996 \pm 0.0030 & $2.1972 \times 10^{-02} \pm 2.5705 \times 10^{-02}$ & $1.7599 \times 10^{-02} \pm 6.9873 \times 10^{-03}$ \\
        & Scheduled Drop & 57.82 \pm 1.33 & 38.21 \pm 0.99 & 0.2976 \pm 0.0082 & $1.6301 \times 10^{-02} \pm 1.7914 \times 10^{-02}$ & $2.4043 \times 10^{-11} \pm 2.0662 \times 10^{-11}$ \\
        & Stochastic Depth & 60.50 \pm 0.29 & 40.16 \pm 0.66 & 0.2703 \pm 0.0005 & $5.4342 \times 10^{-02} \pm 5.0990 \times 10^{-02}$ & $9.8176 \times 10^{-11} \pm 5.7824 \times 10^{-11}$ \\
        & Weight Decay & 57.32 \pm 1.38 & 37.96 \pm 1.48 & 0.3018 \pm 0.0071 & $9.1484 \times 10^{-03} \pm 1.1856 \times 10^{-02}$ & $6.5511 \times 10^{-12} \pm 3.2157 \times 10^{-12}$ \\
        \midrule
        \multirow{12}{*}{ResNet-18} 
        & Baseline & 72.37 \pm 0.72 & 48.36 \pm 4.14 & 0.1977 \pm 0.0037 & $1.0278 \times 10^{+01} \pm 1.7300 \times 10^{+01}$ & $8.3905 \times 10^{+01} \pm 1.4533 \times 10^{-01}$ \\
        & DropConnect & 31.99 \pm 29.20 & 26.68 \pm 24.57 & 0.3050 \pm 0.2158 & $4.4670 \times 10^{-01} \pm 8.0806 \times 10^{-01}$ & $1.1637 \times 10^{+01} \pm 1.0594 \times 10^{+02}$ \\
        & Dropout & 71.61 \pm 2.03 & 48.34 \pm 4.35 & 0.1941 \pm 0.0071 & $1.0858 \times 10^{+01} \pm 9.7259 \times 10^{+00}$ & $1.6117 \times 10^{+02} \pm 3.0594 \times 10^{-02}$ \\
        & LCL & 73.66 \pm 1.71 & 45.47 \pm 1.64 & 0.1912 \pm 0.0127 & $3.6664 \times 10^{-01} \pm 2.2816 \times 10^{-01}$ & $1.3910 \times 10^{+00} \pm 1.1327 \times 10^{-01}$ \\
        & LCL-Reinit-He & 71.66 \pm 2.72 & 47.34 \pm 2.68 & 0.1602 \pm 0.0052 & $3.8145 \times 10^{+01} \pm 5.4406 \times 10^{+01}$ & $2.5190 \times 10^{+00} \pm 2.6940 \times 10^{-01}$ \\
        & LCL-Reinit-He-no-warmup & 71.39 \pm 1.63 & 47.56 \pm 3.37 & 0.1336 \pm 0.0080 & $1.6174 \times 10^{+01} \pm 2.2813 \times 10^{+01}$ & $3.9836 \times 10^{+00} \pm 4.6940 \times 10^{-01}$ \\
        & LCL-Reinit-Xavier & 71.34 \pm 1.20 & 47.19 \pm 3.84 & 0.1733 \pm 0.0114 & $1.1861 \times 10^{+01} \pm 8.0830 \times 10^{+01}$ & $5.3008 \times 10^{+00} \pm 1.2632 \times 10^{+00}$ \\
        & LCL-Reinit-Xavier-no-warmup & 71.14 \pm 1.68 & 47.35 \pm 4.24 & 0.1722 \pm 0.0089 & $2.1039 \times 10^{+01} \pm 2.0787 \times 10^{+01}$ & $1.9828 \times 10^{+00} \pm 2.6232 \times 10^{+00}$ \\
        & LCL-no-warmup & 71.78 \pm 2.83 & 48.10 \pm 2.41 & 0.1951 \pm 0.0065 & $3.9850 \times 10^{-01} \pm 5.0949 \times 10^{-01}$ & $2.3778 \times 10^{-01} \pm 3.0159 \times 10^{-01}$ \\
        & Label Smoothing & 74.83 \pm 2.24 & 51.51 \pm 2.51 & 0.0870 \pm 0.0331 & $1.9374 \times 10^{+00} \pm 4.4046 \times 10^{-01}$ & $8.9299 \times 10^{+00} \pm 9.7109 \times 10^{-01}$ \\
        & RigL & 69.80 \pm 3.59 & 46.30 \pm 0.81 & 0.2169 \pm 0.0191 & $2.9351 \times 10^{+00} \pm 3.3243 \times 10^{+00}$ & $1.3993 \times 10^{+01} \pm 2.3572 \times 10^{+01}$ \\
        & Scheduled Drop & 70.70 \pm 1.97 & 47.13 \pm 4.09 & 0.2034 \pm 0.0055 & $3.5669 \times 10^{+01} \pm 4.0958 \times 10^{+01}$ & $5.6291 \times 10^{+01} \pm 1.6187 \times 10^{+02}$ \\
        & Stochastic Depth & 72.08 \pm 1.61 & 47.60 \pm 4.02 & 0.1863 \pm 0.0023 & $8.4402 \times 10^{-01} \pm 3.3872 \times 10^{-01}$ & $1.1387 \times 10^{+02} \pm 5.2736 \times 10^{+01}$ \\
        & Weight Decay & 72.39 \pm 1.72 & 48.61 \pm 3.65 & 0.1985 \pm 0.0066 & $8.2750 \times 10^{+00} \pm 5.5200 \times 10^{+00}$ & $1.0977 \times 10^{+01} \pm 7.7622 \times 10^{+00}$ \\
        \bottomrule
    \end{tabular}%
    }
\end{table}
For large-scale evaluation, I performed linear probing on the ImageNet-1k dataset~\citep{ILSVRC15}. For this, I used powerful pretrained models, including Swin Transformer~\citep{liu2021Swin}, a model pre-trained on ImageNet-22k~\citep{rw2019timm}; MambaOut-Base-Plus~\citep{yu2024mambaout}, a model pre-trained on ImageNet12k~\citep{rw2019timm}; and EfficientNetV2-M~\citep{tan2021efficientnetv2}, a model pre-trained on ImageNet21k~\citep{rw2019timm}. I froze their backbones and trained only the final classification layer. The final results are shown in Table~\ref{tab:in1k}.

\begin{table}[htbp]
    \centering
    \caption{Performance comparison of various methods under different architectures on ImageNet1k}
    \label{tab:in1k}
    
    \sisetup{
        separate-uncertainty, 
        table-align-text-post=false 
    }
    
    \resizebox{\textwidth}{!}{%
    \begin{tabular}{
        l
        l
        S[table-format=2.2(2)]
        S[table-format=1.4(4)]
        r 
        r 
    }
        \toprule
        \textbf{Model} & \textbf{Method} & {\textbf{ACCURACY (\%)}} & {\textbf{ECE}} & {\textbf{Hessian top eigenvalue}} & {\textbf{Lipschitz bound}} \\
        \midrule
        \multirow{6}{*}{EfficientNetV2-M} 
        & Baseline & 79.21 \pm 0.25 & 0.0653 \pm 0.0331 & $1.1966 \times 10^{+00} \pm 4.1874 \times 10^{-01}$ & $1.1555 \times 10^{-69} \pm 2.9840 \times 10^{-70}$ \\
        & Drop Connect & 79.88 \pm 0.21 & 0.0901 \pm 0.0229 & $1.1788 \times 10^{+00} \pm 9.9937 \times 10^{-01}$ & $1.6893 \times 10^{-69} \pm 6.2454 \times 10^{-70}$ \\
        & Dropout & 80.60 \pm 0.11 & 0.0673 \pm 0.0005 & $7.0280 \times 10^{-01} \pm 3.5335 \times 10^{-01}$ & $1.2979 \times 10^{-69} \pm 8.0208 \times 10^{-73}$ \\
        & LCL & 79.47 \pm 0.16 & 0.0267 \pm 0.0026 & $7.3600 \times 10^{-01} \pm 3.7410 \times 10^{-01}$ & $4.5327 \times 10^{-71} \pm 0.0000 \times 10^{+00}$ \\
        & Label Smoothing & 80.51 \pm 0.05 & 0.1152 \pm 0.0049 & $3.1510 \times 10^{-01} \pm 3.4629 \times 10^{-02}$ & $7.3717 \times 10^{-70} \pm 1.2501 \times 10^{-72}$ \\
        & Weight Decay & 79.14 \pm 0.39 & 0.0662 \pm 0.0346 & $1.6910 \times 10^{+00} \pm 7.8445 \times 10^{-01}$ & $1.1547 \times 10^{-69} \pm 3.0004 \times 10^{-70}$ \\
        \midrule
        \multirow{6}{*}{MambaOut-Base-Plus} 
        & Baseline & 84.76 \pm 0.03 & 0.0272 \pm 0.0006 & $6.6450 \times 10^{-06} \pm 4.1953 \times 10^{-06}$ & $1.6666 \times 10^{-06} \pm 3.6373 \times 10^{-09}$ \\
        & Drop Connect & 84.77 \pm 0.21 & 0.0379 \pm 0.0008 & $5.3900 \times 10^{-06} \pm 3.2147 \times 10^{-06}$ & $1.6364 \times 10^{-06} \pm 1.3299 \times 10^{-08}$ \\
        & Dropout & 84.84 \pm 0.05 & 0.0281 \pm 0.0003 & $7.6089 \times 10^{-04} \pm 1.3173 \times 10^{-03}$ & $1.6719 \times 10^{-06} \pm 3.2747 \times 10^{-09}$ \\
        & LCL & 84.92 \pm 0.08 & 0.0445 \pm 0.0097 & $1.4567 \times 10^{-04} \pm 2.0020 \times 10^{-04}$ & $5.0834 \times 10^{-08} \pm 2.8868 \times 10^{-12}$ \\
        & Label Smoothing & 84.78 \pm 0.05 & 0.2072 \pm 0.0006 & $6.0973 \times 10^{-05} \pm 4.2399 \times 10^{-05}$ & $1.4985 \times 10^{-06} \pm 7.5498 \times 10^{-10}$ \\
        & Weight Decay & 84.83 \pm 0.01 & 0.0276 \pm 0.0003 & $2.1959 \times 10^{-05} \pm 1.5427 \times 10^{-05}$ & $1.6736 \times 10^{-06} \pm 1.6166 \times 10^{-09}$ \\
        \midrule
        \multirow{6}{*}{Swin Transformers Base} 
        & Baseline & 83.32 \pm 0.04 & 0.0199 \pm 0.0002 & $2.3857 \times 10^{-05} \pm 3.0471 \times 10^{-05}$ & $1.5562 \times 10^{+79} \pm 1.5133 \times 10^{+76}$ \\
        & Drop Connect & 83.23 \pm 0.03 & 0.0699 \pm 0.0137 & $5.3240 \times 10^{-06} \pm 5.4957 \times 10^{-06}$ & $2.3201 \times 10^{+79} \pm 5.1445 \times 10^{+78}$ \\
        & Dropout & 83.18 \pm 0.07 & 0.0203 \pm 0.0003 & $4.4510 \times 10^{-05} \pm 4.3814 \times 10^{-05}$ & $1.5553 \times 10^{+79} \pm 1.2702 \times 10^{+76}$ \\
        & LCL & 83.58 \pm 0.07 & 0.0204 \pm 0.0002 & $1.6853 \times 10^{-04} \pm 2.8892 \times 10^{-04}$ & $1.9846 \times 10^{+78} \pm 0.0000 \times 10^{+00}$ \\
        & Label Smoothing & 82.91 \pm 0.05 & 0.0869 \pm 0.0004 & $1.6841 \times 10^{-04} \pm 1.9370 \times 10^{-04}$ & $1.4611 \times 10^{+79} \pm 5.6226 \times 10^{+76}$ \\
        & Weight Decay & 83.22 \pm 0.07 & 0.0208 \pm 0.0005 & $2.7792 \times 10^{-06} \pm 4.5940 \times 10^{-06}$ & $1.5557 \times 10^{+79} \pm 3.6116 \times 10^{+76}$ \\
        \bottomrule
    \end{tabular}%
    }
\end{table}

\FloatBarrier

\section{Theoretical Analysis}
\label{5}
In this section, I provide a theoretical analysis of the Lifecycle (LC) principle. My goal is to explain why this mechanism can act as an effective regularizer. I formalize the training objective under the LC principle. Then, I show that this process favors solutions with lower curvature and reduced model capacity. My analysis directly connects to the empirical metrics I will present in the experiments section.

\subsection{Formalizing the Training Objective}
\label{5.1}
I view the LC principle as a process of applying a stochastic, memory-aware mask to the network's parameters. For a network $f_{\vtheta}$ and a training set $\sS = \{(\vx_i, \vy_i)\}_{i=1}^n$, the standard empirical risk is:

\begin{equation}
\label{eq:loss}
\mathcal{L}(\vtheta) = \frac{1}{n} \sum_{i=1}^n \ell\big(f_{\vtheta}(\vx_i), \vy_i\big).
\end{equation}

For a Lifecycle Linear layer, the forward pass uses a dynamic mask vector $\vd_t \in [0,1]^m$, where $m$ is the number of output neurons. The output is equivalent to
\begin{equation}
\label{eq:lifecycle_forward}
\vy = (\mW \odot \mD_t) \vx + (\vb \odot \vd_t), \qquad \text{where } \mD_t = \mathrm{diag}(\vd_t),\ \vd_t\in[0,1]^{m}.
\end{equation}
The vector $\vd_t$ is determined by the lifecycle process. It is 0 for inactive neurons, 1 for fully active neurons, and between (0,1) for neurons in their warm-up phase.

Let $\mM_t$ represent the collective mask for all LC layers at step $t$. The training process can be seen as optimizing the parameters $\vtheta$ under this sequence of random masks. I define the smoothed training objective $\bar{\mathcal{L}}(\vtheta)$ as the expected loss over the stationary distribution $\mathcal{P}$ of the masks:
\begin{equation}
\label{eq:smoothed_objective}
\bar{\mathcal{L}}(\vtheta) = \mathbb{E}_{\mM \sim \mathcal{P}}\left[\frac{1}{n} \sum_{i=1}^n \ell\big(f_{\vtheta \odot \mM}(\vx_i), \vy_i\big)\right].
\end{equation}
This smoothed objective represents the average loss over a family of subnetworks generated by the lifecycle process. The state memory component does not change this objective. Instead, it stabilizes the optimization path towards a minimum of this objective, which I confirm in my ablation studies.

\subsection{Conclusion A: Smoothing the Loss Landscape and Favoring Flat Minima}
\label{5.2}
The LC mechanism optimizes a smoothed version of the original loss function. This smoothing has a direct effect on the curvature of the loss landscape.
\begin{proposition}
\label{prop:1}
According to the Equation~\ref{eq:smoothed_objective}, the Hessian of the smoothed objective $\bar{\mathcal{L}}(\vtheta)$ is an expectation over the Hessians of the masked objectives. Assume that $\ell$ is quadratically differentiable and that the exchange of expectation and derivative is legal (which is valid under finite samples and bounded gradients). Then:
\[
\nabla^2_{\vtheta}\bar{\mathcal{L}}(\vtheta) = \mathbb{E}_{\mM \sim \mathcal{P}}\left[ \mathrm{Diag}(\mM) \nabla^2_{\vtheta} \mathcal{L}(\vtheta \odot \mM) \mathrm{Diag}(\mM) \right].
\]
\end{proposition}
This means the masking process attenuates the curvature. The maximum eigenvalue of the smoothed Hessian is bounded by the maximum eigenvalue of the original Hessian. This suggests that the optimizer is guided towards flatter minima. Prior work has established a strong connection between flat minima and better generalization performance~\citep{keskar2017on}.

\begin{proposition}
\label{prop:2}
In the neighborhood of $\vtheta$, let the mask be $\mM = \mathbf{1} - \mathbf{\xi}$, where $\mathbf{\xi} \in [0,1]^d$ indicates a slight weakening of the multiplication. We perform a second-order Taylor expansion on $\mathcal{L}(\vtheta \odot \mM) = \mathcal{L}(\vtheta \odot (\mathbf{1} - \mathbf{\xi}))$ and take the expectation. If the gradient vanishes at $\vtheta$ (i.e., $\nabla_{\vtheta} \mathcal{L}(\vtheta) = 0$), then the smoothed objective is approximately:
\[
\bar{\mathcal{L}}(\vtheta) \approx \mathcal{L}(\vtheta) + \frac{1}{2} \sum_{i=1}^d \mathrm{Var}[\mathbf{\xi}_i]\vtheta_i^T\left( \nabla^2_{\vtheta} \mathcal{L}(\vtheta) \right)\vtheta_i.
\]
\end{proposition}

The second term on the right is a curvature-weighted "ridge" penalty: the greater the curvature (sharp minima), represented by $\nabla^2_{\vtheta} \mathcal{L}(\vtheta)$, or the larger the coordinate magnitude,$||\vtheta||^2$, the more $\bar{\mathcal{L}}$ penalizes it. Unlike ordinary ridge regression, $\lambda\|\vtheta\|_2^2$, this penalty is adaptively weighted by the Hessian, forming a "curvature-aware ridge".

I empirically validate this proposition. I use the top eigenvalue of the Hessian matrix to measure the sharpness of the final solution and use power iteration to estimate this eigenvalue. My theory predicts that the LCL-trained model exhibits significantly lower top eigenvalues of the Hessian matrix compared to the baseline model. This is demonstrated in Tables~\ref{tab:cifar10},~\ref{tab:cifar100} and~\ref{tab:in1k}.

\subsection{Conclusion B: Reducing Model Capacity for Better Generalization}

The LC mechanism also acts as a form of capacity control. It reduces the expected complexity of the function class that the network can represent. This can be analyzed through the network's Lipschitz constant.

\begin{proposition}
\label{prop:3}
The expected Lipschitz constant of a network with LC layers is contracted. For an L-layer network with 1-Lipschitz activations, the expected Lipschitz constant is bounded by:
\[
\mathbb{E}_{\mM}\big[\mathrm{Lip}(f_{\vtheta \odot \mM})\big] \le \left(\prod_{\ell=1}^L \mu_\ell\right) \cdot \prod_{\ell=1}^L \|\mW^{(\ell)}\|_2.
\]
\end{proposition}
Here, $\|\mW^{(\ell)}\|_2$ is the spectral norm of the weight matrix of layer $\ell$. The factor $\mu_\ell = \sqrt{\mathbb{E}[(\vd^{(\ell)})^2]} \in (0,1]$ is less than 1 because neurons are inactive or warming up for a fraction of the time. A smaller Lipschitz constant implies a smaller Rademacher complexity of the function class~\citep{bartlett2002rademacher}. This leads to a tighter generalization bound.

Let $\mathcal{F}_{\text{base}}$ be the class of functions without LCL, and $\mathcal{F}_{\text{LCL}}$ be the class of functions with random masking (independent of the data), then the empirical Rademacher complexity satisfies:
\begin{equation}
\label{eq:rademacher_bound}
\hat{\mathfrak R}_n(\mathcal{F}_{\text{LCL}}) \lesssim \left(\prod_{\ell=1}^L \mu_\ell\right) \cdot \hat{\mathfrak R}_n(\mathcal{F}_{\text{base}}),
\end{equation}
Thus, the generalization error bound is tightened:
\begin{equation}
\label{eq:gen_gap_bound}
\mathbb{E}[\text{gen\_gap}] \lesssim
\tilde{O}\!\left( \left(\prod_{\ell=1}^L \mu_\ell\right) \cdot \frac{ \prod_{\ell=1}^L \|\mW^{(\ell)}\|_2 \cdot \mathrm{diam}(\mathcal X)}{\sqrt n} \right).
\end{equation}
That is, the greater the masking strength (the smaller $\mu_\ell$), the tighter the generalization upper bound.

I provide evidence for this capacity reduction. I compute a proxy for the network's Lipschitz constant by multiplying the spectral norms of its layers' weight matrices. My theory predicts that the LCL-trained model achieves strong performance while maintaining a smaller Lipschitz constant proxy compared to the baseline. This is demonstrated in Tables~\ref{tab:cifar10},~\ref{tab:cifar100} and~\ref{tab:in1k}.

\subsection{Conclusion C: Reducing Co-adaptation}

\begin{proposition}
\label{prop:4}
Under the first-order necessary conditions of $\bar{\mathcal{L}}$,
\[
\mathbb E_{\mM}\!\big[\mM \odot \nabla_{\vtheta} \mathcal{L}(\vtheta\odot \mM)\big]=0.
\]
\end{proposition}
Because $\mathbb P[d_i=0]>0$ for each coordinate $i$, this condition forces the overall target to remain stationary even when any single neuron is absent with a certain probability—this is equivalent to weakening the reliance on individual neurons/connections. This is similar to the "ensemble effect" of dropout, but LCL is stronger: the absence is "long-windowed" (lifetime length) rather than instantaneous and random on a sample-by-sample basis, thus more substantially suppressing co-adaptation.

\begin{proposition}
\label{prop:5}
Assume that $\mathcal{L}$ has a large Hessian eigenvalue in the direction of the principal curvature of a sharp minimum $\vtheta^\sharp$. If $\mathrm{Var}[d_i]>0$ and the warm-up is such that $d_i\in(0,1)$ within the non-ignored fraction of training, then according to Proposition~\ref{prop:2}, under the second-order approximation of $\vtheta^\sharp$, the additional term introduced by $\bar{\mathcal{L}}$ in this direction is $\tfrac12 \mathrm{Var}[d_i]\cdot \vtheta_i^\top H(\vtheta^\sharp)\vtheta_i$, and its magnitude increases linearly with the curvature. Therefore, $\vtheta^\sharp$ is often not a minimum of $\bar{\mathcal{L}}$ (or its basin becomes significantly shallower/narrower). Under LCL, the scheduler is less likely to be "locked" by the sharp basin and tends to shift to flatter regions.
\end{proposition}

The reduction in co-adaptation leads to improved robustness. I verify this by evaluating the final model's performance on corrupted test sets, such as CIFAR-10-C~\citep{hendrycks2018benchmarking}. My theory predicts that the LCL-trained model demonstrates higher accuracy on these datasets, indicating that it has learned more robust features.

\section{Conclusion}
\label{6}
In this paper, I introduced the Lifecycle (LC) principle, a new regularization mechanism for training neural networks. The core idea is to let individual neurons undergo long-term cycles of deactivation and revival. I identified that this dynamic process can cause training instability. To address this, I proposed a state memory component as the key part of my method. This component allows a revived neuron to restore its previously learned parameters, which effectively mitigates optimization shocks.

My theoretical analysis suggests that the LC principle smooths the loss landscape and encourages the optimizer to find flatter minima. My experiments show that my implementation, Lifecycle Linear, improves generalization and robustness over standard baselines. The results also confirm that state memory is critical for the method's success.

The Lifecycle principle is a general concept. In this work, I only applied it to fully connected layers. An important direction for future research is to extend this principle to other architectures. For example, one could apply it to the channels of convolutional neural networks or to the attention heads in Transformers~\citep{vaswani2017attention}. Exploring how the Lifecycle principle interacts with different model types is a promising area for future work.

\bibliography{iclr2026_conference}
\bibliographystyle{iclr2026_conference}

\appendix
\section{Algorithm}
\begin{algorithm}[h]
\caption{Lifecycle Linear Update Step}
\label{alg:lifecycle-linear-final}
\begin{algorithmic}[1]
\Require
    Weights $W \in \mathbb{R}^{m \times d}$, Bias $b \in \mathbb{R}^{m}$; 
    Memory buffers $W^{\text{mem}} \in \mathbb{R}^{m \times d}$, $b^{\text{mem}} \in \mathbb{R}^{m}$;
    State buffers: activity mask $a \in \{0,1\}^{m}$, warm-up scaler $s \in [0,1]^{m}$, life counters $c^{\text{life}} \in \mathbb{R}^{m}$, recovery counters $c^{\text{rec}} \in \mathbb{R}^{m}$;
    Hyperparameters: max durations $L_{\max} \in \mathbb{R}^{m}$, $R_{\max} \in \mathbb{R}^{m}$, warm-up steps $k \in \mathbb{N}_{+}$.
\Statex
\Procedure{Initialize}{}
    \State $a \gets \mathbf{1}$; \quad $s \gets \mathbf{1}$; \quad $c^{\text{rec}} \gets \mathbf{0}$
    \State $c^{\text{life}}_i \sim U[l_{\min}, l_{\max}]$ for $i=1, \dots, m$
    \State $W^{\text{mem}} \gets W$; \quad $b^{\text{mem}} \gets b$
\EndProcedure
\Statex
\If{not in training mode} \Return \Comment{Lifecycle updates only occur during training.}
\EndIf
\Statex

\AlgComment{\textbf{(A) Warm-up Phase: Ramp up recently revived neurons}}
\State $I_{\text{warm-up}} \gets \{i \mid s_i < 1\}$ \Comment{Perfectly matches the Python code.}
\State $s_{I_{\text{warm-up}}} \gets \min(1, s_{I_{\text{warm-up}}} + 1/k)$
\Statex

\AlgComment{\textbf{(B) Update Timers}}
\State $c^{\text{life}} \gets c^{\text{life}} - a$
\State $c^{\text{rec}} \gets c^{\text{rec}} - (1 - a)$
\Statex

\AlgComment{\textbf{(C) Lysis Phase: Deactivate expired neurons}}
\State $I_{\text{lyse}} \gets \{i \mid a_i = 1 \land c^{\text{life}}_i \le 0\}$
\If{$I_{\text{lyse}} \neq \emptyset$}
    \State $W^{\text{mem}}[I_{\text{lyse}},:] \gets W[I_{\text{lyse}},:]$ \Comment{Save parameters to memory.}
    \State $b^{\text{mem}}[I_{\text{lyse}}] \gets b[I_{\text{lyse}}]$
    \State $a_{I_{\text{lyse}}} \gets 0$
    \State $c^{\text{rec}}_{I_{\text{lyse}}} \gets R_{\max, I_{\text{lyse}}}$
    \State $s_{I_{\text{lyse}}} \gets 1$ \Comment{Reset scaler for the next revival.}
\EndIf
\Statex

\AlgComment{\textbf{(D) Revival Phase: Reactivate recovered neurons}}
\State $I_{\text{revive}} \gets \{i \mid a_i = 0 \land c^{\text{rec}}_i \le 0\}$
\If{$I_{\text{revive}} \neq \emptyset$}
    \State $W[I_{\text{revive}},:] \gets W^{\text{mem}}[I_{\text{revive}},:]$ \Comment{Restore from memory.}
    \State $b[I_{\text{revive}}] \gets b^{\text{mem}}[I_{\text{revive}}]$
    \State $a_{I_{\text{revive}}} \gets 1$
    \State $c^{\text{life}}_{I_{\text{revive}}} \gets L_{\max,I_{\text{revive}}}$
    \State $s_{I_{\text{revive}}} \gets 1/k$ \Comment{Initiate warm-up.}
\EndIf
\end{algorithmic}
\end{algorithm}

\FloatBarrier

\section{Hyperparameters}
\begin{table}[h]
\centering
\caption{Key hyperparameters in CIFAR-10 and CIFAR-100 training configurations.}
\label{tab:main_hyperparams_cifar}
\begin{tabular}{@{}lll@{}}
\toprule
\textbf{Category} & \textbf{Parameter} & \textbf{Value} \\ \midrule
\textbf{Dataset} & Validation Split & 10\% of the training data \\ \midrule
\textbf{Training Dynamics} & Epochs & 100 \\
 & Batch Size & 128 \\
 & Loss Function & Cross-Entropy Loss (except for the Label Smoothing experiment) \\
 & Random Seeds & 42, 123, 888 (3 separate runs) \\ \midrule
\textbf{Optimizer} & Optimizer & AdamW ($\beta_1=0.9, \beta_2=0.999$) \\
 & Learning Rate (LR) & $1 \times 10^{-3}$ \\
 & LR Scheduler & Warmup Cosine Decay \\
 & Warmup Epochs & 5 \\ \midrule
\textbf{Data Augmentation} & Transformations & RandomCrop, RandomHorizontalFlip \\
 & Normalization & The mean and standard deviation of the data set \\ \bottomrule
\end{tabular}
\end{table}

\begin{table}[h]
\centering
\caption{Key hyperparameters in ImageNet-1k training configurations.}
\label{tab:main_hyperparams_in1k}
\begin{tabular}{@{}lll@{}}
\toprule
\textbf{Category} & \textbf{Parameter} & \textbf{Value} \\ \midrule
\textbf{Dataset} & Validation Split & 10\% of the training data \\ \midrule
\textbf{Training Dynamics} & Epochs & 15 \\
 & Batch Size & 256 \\
 & accumulation & 32 \\
 & Loss Function & Cross-Entropy Loss (except for the Label Smoothing experiment) \\
 & Random Seeds & 42, 123, 888 (3 separate runs) \\ \midrule
\textbf{Optimizer} & Optimizer & AdamW ($\beta_1=0.9, \beta_2=0.999$) \\
 & Learning Rate (LR) & $1 \times 10^{-2}$ \\
 & LR Scheduler & Warmup Cosine Decay \\
 & Warmup Epochs & 5 \\ \midrule
\textbf{Data Augmentation} & Transformations & RandomCrop, RandomHorizontalFlip \\
 & Normalization & The mean and standard deviation of the data set \\ \bottomrule
\end{tabular}
\end{table}

\end{document}

%% file: math_commands.tex
\usepackage{amsmath,amsfonts,bm}

\def\eqref#1{equation~\ref{#1}}

\def\1{\bm{1}}

\def\vtheta{{\bm{\theta}}}

\def\vb{{\bm{b}}}

\def\vd{{\bm{d}}}

\def\vx{{\bm{x}}}
\def\vy{{\bm{y}}}

\def\mD{{\bm{D}}}

\def\mM{{\bm{M}}}

\def\mW{{\bm{W}}}

\DeclareMathAlphabet{\mathsfit}{\encodingdefault}{\sfdefault}{m}{sl}
\SetMathAlphabet{\mathsfit}{bold}{\encodingdefault}{\sfdefault}{bx}{n}

\def\sS{{\mathbb{S}}}

